# Psychological and Normative Theories of Causal Power and the Probabilities of Causes

Clark Glymour
Department of Philosophy
Carnegie Mellon University
University of California, San Diego
cg09@andrew.cmu.edu

## Abstract

This paper (1) shows that the best supported current psychological theory (Cheng, 1997) of how human subjects judge the causal power or influence of variations in presence or absence of one feature on another, given data on their covariation, tacitly uses a Bayes network which is either a noisy *or* gate (for causes that promote the effect) or a noisy *and* gate (for causes that inhibit the effect); (2) generalizes Cheng's theory to arbitrary acyclic networks of noisy or and noisy and gates; (3) gives various sufficient conditions for the estimation of the parameters in such networks when there are independent, unobserved causes; (4) distinguishes direct causal influence of one feature on another (influence along a path with one edge) from total influence (influence along all paths from one variable to another) and gives sufficient conditions for estimating each when there are unobserved causes of the outcome variable; (5) describes the relation between Cheng models and a simplified version of the "Rubin" framework for representing causal relations.

## I. Cheng's Model of Human Judgement of Causal Power.

Consider variables, A, B, C, D, E, etc., each taking values 1 (present) or 0 (absent). Given data on the joint frequency of candidate causes (of effect E) and of E, when unobserved causes of E may also be acting, how do people judge the efficacy or causal power of any particular observed candidate cause? Cheng (1997) proposes that, according to context, people have implicit models of the causal relations among the variables, and in appropriate circumstances their judgements of causal power are at least qualitatively correct in view of those models. Cheng's theory postulates that the human model of facilitating causes--those which increase the probability of the effect--is distinct from the human model of preventive causes--those which decrease the probability of the effect. She deals explicitly with two cases.

Case 1: All unobserved causes of E are facilitating, and one or more facilitating candidate causes of E are observed along with E. For the simplest case in which C is the only observed facilitating candidate cause and U is an unobserved facilitating cause she proposes:

(1) $Pr(E = 1) = Pr(q_{e,c}C = 1) + Pr(q_{e,u}U = 1) - Pr(q_{e,c} q_{e,u}CU = 1)$

The q parameters are assumed to be independent of each other and of the variables. It follows that

(2) $Pr(E = 1 \mid C = 1, U = 0) = Pr(q_{e,c} = 1)$.

which justifies describing $Pr(q_{e,c} = 1)$ as the "causal power" of C to produce E. Cheng shows that when C, U are independent:

(3) $Pr(q_{e,c} = 1) = [Pr(E = 1 \mid C = 1) - Pr(E = 1 \mid C = 0)] / [1 - Pr(E = 1 \mid C = 0)]$

So that the causal power of C to produce E can be estimated without observing U. Hereafter the numerator in the r.h.s. of (3) will be denoted by $\Delta P_C$.

When two or more independent facilitating causes are observed the causal power of any one of them, say C, can be obtained by applying the same formulas (1), (2), (3). When two or more correlated candidate facilitating causes, say C and D, are observed, Cheng obtains the causal power of any one of them, say C, by using conditional probabilities (on D = 0) in (1), (2) and (3).

Case 2: All unobserved causes U of E are facilitating, and there is an observed candidate preventing cause F of E. Cheng's equation is:

(4) $Pr(E = 1) = Pr(q_{e,u} U = 1) \cdot Pr(1 - q_{e,f} F = 1)$.

Assuming independence of U and F she shows that

(5) $Pr(q_{e,f} = 1) = - \Delta P_f / Pr(E | F = 0)$

(5) shows that the parameter $Pr(q_{e,f} =1)$ can be estimated from observations of F, E alone, even in the presence of other unrecorded facilitating causes of E, so long as they are independent of F. Cheng proposes that when, in such circumstances, subjects report their estimates of the power of F to prevent E, they are reporting an estimate of $Pr(q_{e,f}=1)$. Her account of judgements of facilitating and preventing causes is justified by an extensive review of the literature in experimental psychology on human judgement of causal power.

## II. Issues about Cheng's Theory

Several questions naturally arise about implications of the model: (1) What happens if, as in cases 1 and 2, the unobserved causes are facilitating, but the observed causes include both facilitating and preventing causes? Can the causal powers of both be estimated? (2) What if all observed causes are preventing, but more than one is observed? (3) What if some or all of the unobserved causes are preventing? (4) What if a cause, whether facilitating or preventing, only influences the effect indirectly, through intervening observed causes, either facilitating or preventing? (5) What if there are multiple pathways, some through observed intervening causes, from a cause to the effect? (6) What if a preventing causes interferes with some facilitating causes but not others? (7) What if a cause prevents some effects and facilitates others? (8) Or consider a case like 1, but with two facilitating observed correlated candidate causes, C and D, for which D is the unique facilitating cause of C. To determine the causal power of C by the method described, probabilities must be estimated conditional on C = 1 and D = 0, which is inconsistent with the model. Can the causal power of C nonetheless be estimated in some other way in such cases? (9) Instead of estimating the probability that a factor C causes E given that C occurs, can we estimate the probability that C caused E, given that C and E both occur? The last question is of enormous practical importance in scientific and legal contexts. (10) Can the model be used to forecast the frequency of the effect if an intervention removed (or prevented) the cause in all units? (11) And, not least, how if at all is Cheng's model related to graphical causal models of the sort widely used in the social sciences and advocated by Spirtes, et al., 1993 and by Pearl, 1995, and to the representation of causation now widely used in statistics advocated by Rubin, Holland (1985) and others?

Complete or partial answers to all of these questions about Cheng models can be found by recognizing them as particular parameterizations of Bayes nets. The arguments require only algebra and some well known ideas about directed graphical models; proofs will not be given here, but they are all straightforward. An interesting, and in view of Cheng's empirical results perhaps not accidental, historical fact is that directed graphical models seem first to have been applied in computer science because they facilitated elicitation of expert judgements for use in probabilistic expert systems. While I hope the results that follow are interesting in themselves, they illustrate the wonderful power of directed graphical formulations to extract implications from other representations.

## III. The Quickest Review of Bayes Nets with a Causal Interpretation.

Where C, D, E etc. are causally related represent their causal relations by drawing appropriate directed edges among the variables, for example:

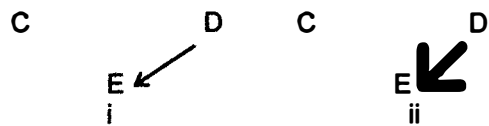

i        ii

When drawing such graphs, a preventing influence will be written as a boldface directed arrow; unobserved variables will be written inside rectangles.

Let G be a directed acyclic graph (hereafter DAG; acyclic means no directed paths (see below) from a vertex back to itself) and let Pr be a probability distribution defined on a set **S** of vectors of values of the variables represented by vertices in G. Then we say Pr is *Markov* for G, **S** if, for all vectors X of values in **S**, Pr(X) equals the product, over all variables, of the probability (by Pr) of the value of each variable conditional on the values of its parents. Any such decomposition for values of variables represented by vertices in a DAG will be called a Markov factorization of Pr. When I write equations in terms of variables, as in Pr(C,D,E) = Pr(E | C,D) Pr(C) Pr(D) I mean the analogous equation holds for all permissible values of the variables.

The *indegree* of a vertex is the number of edges directed into it. A directed path from vertex C to vertex E in a DAG is a sequence of vertices whose first member is C and whose last is E such that every member in the sequence has an edge directed into it from the vertex immediately preceding it in the sequence.

Let v be a value for a variable V represented by a vertex V in a DAG G, with admissible value set **S**, and associated Markov probability distribution Pr. A V(v) *intervention* is a graph G(V=v) obtained from G by removing all edges directed into V, and an admissible value set **S\*** obtained from **S** by replacing with v any occurrence of a value v'_ v of V in a member of **S**, and a probability distribution Pr* obtained by replacing, in each Markov factorization of Pr, all factors of the form Pr(V = v' | ___) by 1, and all factors of the form Pr (V = v' | ___) by zero if v' ≠ v, and by replacing Pr( __ | V = v') by Pr( __ | V = v) if defined, and if not defined then by some new distribution Pr*( __ | V = v). An example, illustrating the need for this complexity, will be given in section **V**.

## IV. Answering Half of Question 11: Cheng Models are Noisy Or and Noisy And Gates, and so Causal Bayes Nets

Equation 1 can be obtained by taking probabilities on both sides of a deterministic equation relating Boolean variables, and using Cheng's assumption that the q parameters are independently distributed.

(6) $E = q_{e,c}C ++ q_{e,u}U$

where "++" denotes Boolean addition. Probabilities based on this equation, the independence of the parameters, and the further independence of C and U are *noisy or gates* (Pearl, 1988). They can be represent by a particular parameterization of probability distributions Markov for a DAG, implying the independence of C and U. Adding additional sums to equation (6) and taking probabilities, the number of independent facilitating causes can be increased arbitrarily. Equation 5 can similarly be obtained by taking probabilities in the equation

(7) $E = q_{e,c}C \cdot (1 - q_{e,f} F)$

With C and F independent, this represents a noisy *and* gate. Clearly any composition of facilitating and preventing causes can be represented by a system of such equations.

For any DAG, G, with edges labeled as facilitating or preventive, and, if necessary, preventive edges labeled for the facilitating edges they collide with and interfere with, there is a corresponding set of Boolean equations. Let S be the set of values of variables consistent with the equations. Let Pr be positive on **S**, and Markov for G . I will call any such structure, < G, **S**, Pr>, such that every vertex of positive indegree has at least one facilitating cause, *a Cheng model*. (Variables whose only causes are preventing are fixed at zero and so such structures will be ignored.)

## V. Answering Question 8: Cheng Models with Interventions

For equations and DAG

(8) $E = q_{e,c} C + q_{e,d} D$

(9) $D = q_{d,c} C$

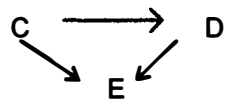

Pr($q_{e,d}$) cannot be computed from Pr(C,D,E). A computation using equation (3) with probabilities conditional on C = 0 would require Pr(E | C = 0,D = 1), which is undefined, since D must vanish when C does. Pr($q_{e,d}$) can, however, be calculated

from the (1,D) intervention in the structure just given. That intervention yields

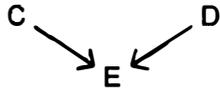

and the equation

(10) $E = q_{e,c} C + q_{e,d}$

and the new probability distribution $Pr^*(C = 0, D = 1, E) = Pr^*(E | D = 1, C = 0)Pr(C = 0)$; $Pr^*(C = 1, D = 1, E) = Pr(E | D = 1, C = 1)Pr(C = 1)$. (The equations resulting from a (v,V) intervention can always be obtained by substituting v for V in any equation in which V occurs on the r.h.s. and eliminating the equation in which V occurs on the l.h.s. See Pearl (1995)).

$Pr^*(q_{e,d})$ can now be immediately calculated by setting $C = 0$ in (10) and taking $Pr^*$ of both sides. Assuming $Pr(q_{e,d}) = Pr^*(q_{e,d})$, the original causal power can be estimated.

The hypothesis that in appropriate situations people implicitly use Cheng models in judging causal power thus implies that *given frequency data on two facilitating causes and their effect, when the value of one of the facilitating causes is uniquely determined by the other, subjects should be uncertain as to the causal power of the determined facilitating cause, but that uncertainty should be reduced if they are afforded the opportunity to intervene and manipulate the values of the determined facilitating cause.*

## VI. Answering Questions 2 and 3 and (Part of) 6: Preventing Causes

Cheng's equation (5) shows that the causal power of a preventing cause can be estimated when there is an independent unobserved facilitating cause. The derivation can be generalized to any number of independent observed preventive causes and an unobserved facilitating cause, although no experiment testing human judgement with more than one observed preventing cause seems to have been published.

The situation is not symmetric. If there is an unobserved cause F, whose occurrence tends to prevent an independently occurring observed facilitating cause C from bringing about E, then the causal power of C, $q_{e,c}$, cannot be estimated from Pr(CAE) or from C interventions.

In appropriate experimental situations, *subjects should be uncertain of the power of a facilitating cause when they have reason to think a preventing cause is acting.*

If, however, there are two facilitating causes, C,D and an independent unobserved cause F which tends to prevent the influence of D on E but not the influence of C on E, then the causal power of C, but not D, can be estimated. The equation is

(11) $E = q_{e,c} C + [q_{e,d} D \cdot (1 - q_{e,f} F)]$

and the derivation of $Pr(q_{e,c})$ is the same as for (3) (see Cheng, 1997) but with the expression in brackets substituted for U. *Subjects using Cheng models should have differential confidence in judgements of causal power when they know there is an unobserved preventing cause of one of two observed facilitating causes.*

## VII. Answering Questions 4 and 5: Direct Versus Indirect Causal Power.

A Cheng model with a DAG

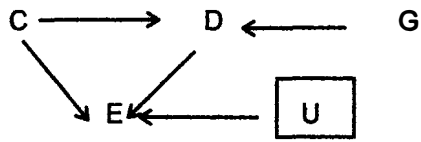

with C facilitating, provides two distinct senses of the causal power of C: the direct causal power of C, which is $Pr(E = 1 | C = 1, D = 0, G = 0, U = 0) = Pr(q_{e,c})$ and the *total* causal power of C, which is $Pr(E = 1 | C = 1, G = 0, U = 0)$. The total causal power of C is the probability that E occurs given that C occurs and that *no other causes of E -- direct or indirect--occur unless they are effects of C.* If D is facilitating as well, then it follows directly from the equations for the model that

(12) $Pr(E = 1 | C = 1, G = 0, U = 0) = Pr(q_{e,c} + q_{d,c} q_{e,d} = 1)$.

For any Cheng model in which all causes are facilitating, the total causal power of any variable, C, to influence any other, E--the probability that E = 1 given that C = 1 and that all causes of E that are not influenced by C have value 0--is the probability of the Boolean sum, over all directed paths from C to E, of the product of the q coefficients associated with the edges in the path. The preceding examples afford a simple illustration of this rule, which is similar to the rule for computing influence in linear path models with standardized variables. For such Cheng models, the total causal power can be computed piecewise by computing $Pr(q_{a,b})$ for each edge B -> A in each directed path from C to E and using the independence of the q coefficients. But the total causal power can be computed more directly by using equation (3) with a probability distribution formed from Pr by conditioning on zero values for all observed causes of E that are not influenced by C.

*In appropriate contexts, and subject to processing limitations, subjects using Cheng models should give distinct judgements for total causal power and direct causal power, and should be able to judge the total causal power of any variable on any other.*

## VIII. More on Question 4: Pathways with Both Facilitating and Preventing Causes

If a cause C influences E only through a preventive direct cause F of E, which C facilitates, then C is not facilitating for E. Thus for the Cheng model.

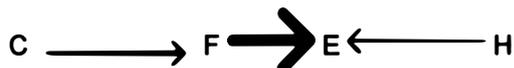

$E = q_{e,h} H \cdot (1 - q_{e,f} F)$

$F = q_{f,c} C$

we have

(13) $E = q_{e,h} H \cdot (1 - q_{e,f} q_{f,c} C)$

and the probability that E = 1 is obviously greater when C = 0 than when C = 1. The appropriate measure of the total causal power of C (to *prevent* E) is thus $Pr(q_{e,f} q_{f,c} = 1)$. In such cases Cheng's formula (5) can be used to identify the total causal power of C.

*Subjects using Cheng models should judge that a cause that facilitates an intermediate cause that prevents an effect, prevents the effect. Subjects should be able to make correct (qualitative) judgements about causal power in such circumstances.*

When a cause C has two or more pathways to an effect E, some pathways with a preventing direct cause of E and some with only facilitating causes, the cause always facilitates the effect.

For example

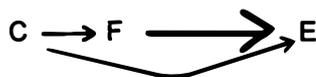

$E = (1 - q_{e,f} F) \cdot q_{e,c} C$

$F = q_{f,c} C$

Hence

$E = C q_{e,d} (1 - q_{e,c} q_{e,f} q_{f,c})$

(14) $P(E = 1 | C = 0) = 0$. $P(E = 1 | C = 1) = Pr(C q_{e,d} (1 - q_{e,c} q_{e,f} q_{f,c}) = 1) > 0$.

If C is a facilitating cause of F, which is a preventing cause of D, which is a facilitating cause of E, then C is a preventing cause of E, because D must have some other cause U, which facilitates D and hence E, but whose influence F reduces. For example:

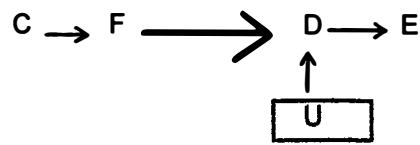

Qualitative reasoning suffices: the probability that F = 1 is greater given C = 1 than given C = 0. The probability that D = 1 is greater given F = 0 than given F = 1, and the probability that E = 1 is greater given that D = 1 than that D = 0, and the probabilities are independent. C has no causal power to produce E.

The measure of causal power for a cause C that, by different pathways, both facilitates and prevents an effect E, is tricky. For example

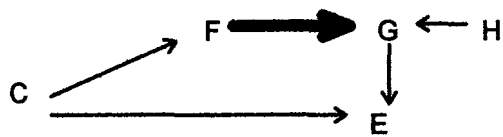

Because F is preventive, G must have some other cause, say H. The route from C to E through F tends to prevent E, while the direct connection facilitates E. The total causal power measured by the probability of E given that C =1 and all other causes of E that are not effects of E is simply $P(q_{e,c})$, because we must condition on H = 0, which implies that G = 0. In general, a path with a preventing cause contributes to the total causal power only if the preventing cause is a parent of E. It would be interesting to know if human judgement is sensitive to these intricacies

The outermost function of a Boolean expression for a Cheng model must be either a product or a sum. If a sum, a generalization of equation (3), conditioning on all observed causes E that are not effects of the candidate cause C, can be applied to compute the total causal power; if a product, then a generalization of equation (5). The direct, and the total causal power of any observed variable C with regard to any other observed variable E can be estimated from frequencies in any Cheng model, provided (i) no observed variable on any pathway from C to E is a deterministic function of any set of other such variables, and (ii) if D is a direct observed facilitating cause of observed cause G on a directed path from C to E, then G has no unobserved cause which prevents the influence of D. Where variables are related deterministically, parameters in Cheng models can still be estimated by interventions, provided condition (ii) applies.

## IX. Answering Questions 10 and 9: Cheng Models in Law and Epidemiology.

The standard formula for predicting the proportion of injuries (E) that would be avoided if a risk factor (C) were eliminated from a population is the Population Attributable Fraction (PAF) (Finkelstein and Levin, 1990):

(15) $\Delta P_C$ Prob(C) / Prob(E)

Assuming only that C is independent of other causes of E and of the q parameters, Cheng's theory implies that the probability that C alone caused E, given that E occurs, equals PAF. Let U represent all other potential causes of E. Then in the Cheng model the frequency of units with E caused by C alone is

(16) Prob($q_{ec}$ • C • (1 - $q_{e,u}$ • U) = 1) =
Prob(C = 1)[Prob($q_{e,c}$ = 1) - Prob($q_{ec}$ = 1)Prob($q_{eu}$ • U = 1)]

The factor in square brackets is easily seen to equal $\Delta P_c$. Dividing by Prob(E) yields PAF.

The quantity relevant to legal disputes in which injury and exposure to a risk factor are stipulated is the probability that the risk factor alone caused the injury, conditional on occurrence of injury and exposure to the risk factor. The standard formula used (Finkelstein and Levin, 1990) is equivalent to:

(17) $\Delta P_C$ Prob(C) / Prob(E,C)

which follows immediately for Cheng models from the derivation of equation (15).

To illustrate, consider the following data (from Finkelstein and Levin, 1990) comparing Leukemia deaths in children in Southern Utah with high and low exposure to radiation from fallout from nuclear tests in Nevada:

|  | High | Low |
|---|---|---|
| Deaths | 30 | 16 |
| Person years (in hundreds) | 6,913 | 5,901 |

For Southern Utah the PAF is .245. The probability that leukemia death is caused by high exposure, given death and high exposure, is .3756.

## X. The Rest of Question 11: Cheng, Cartwright and Rubin Models.

Cartwright (19??) advanced the two ideas, realized in Cheng models, that causal power is a feature of properties and is measured by the probability of the effect property when the causal property obtains and no other causal properties obtain. At least the first idea is implicit in most treatments of causation in the social sciences. Rubin, Holland and others have advanced a different representation of causal relations, according to which the locus of causal power is not in the properties but in the units, which differ in their disposition to elicit the effect property in response to the occurrence of the causal property. The difference is metaphysics, but not just metaphysics . A simple model in the Rubin-Holland spirit is as follows: Every unit in the sample is one of four mutually exclusive kinds:

**kind c**: if exposed to C, then exhibits property E; otherwise does not exhibit E.
**kind u**: if exposed to other causes U of E (besides C), then exhibits property E; otherwise does not exhibit E
**kind cu**: exhibits E if exposed to C or to other causes of E
**kind n**: does not exhibit E no matter what causes (C or U) it is exposed to.

Let the frequencies of these kinds in the sample S be Prob(c); Prob(u); Prob(cu); Prob(n). We assume that these frequencies are all independent of C, which is also independent of U. It follows that:

(18). Prob(E) = Prob(C)Prob(c) + Prob(U)Prob(u) + Prob(U)Prob(cu) + Prob(C)Prob(cu) - Prob(U)Prob(C)Prob(cu).

Equations 15 and 17, and their interpretation, follow straightforwardly from equation 18 and its interpretation. Indeed, equations 15 and 17 follow if the "Rubin" model is extended to allow interaction by positing a fifth kind of unit that responds with E only when C and U are both present. The Cheng model allows a similar extension, under the assumption that C is independent of U and of the q parameters. The simple Cheng model (assuming only that C is independent of U and of the q parameters) and the simple Rubin model are related :

(19) Prob(u) + Prob(cu) = Prob($q_{e,u}$)
(20) Prob(c) + Prob(cu) = Prob($q_{e,c}$)

Adding to the Cheng model the assumption that $q_{e,u}$ and $q_{e,c}$ are independent imposes a constraint that is not satisfied in the "Rubin" model unless:

(21) Prob(u)Prob(c) =
        Prob(cu)
   1 - Prob(u) - Prob(c) - Prob(cu)

In particular, if U is not observed, then the "causal power," Prob(E = 1 | C = 1, U = 0), cannot be estimated in the "Rubin" model unless equation 21 holds (and makes little sense in any case in the Rubin model). Cheng's results may indicate that human judgement of causation is better represented by Cartwright's than by Rubin's view of causation.

### Acknowledgements

I thank Patricia Cheng for many very helpful conversations, and, as usual, I am indebted to Peter Spirtes, who pointed out that Cheng's model of facilitating causes is a noisy or gate. This work was supported by a contract from the Office of Naval Research and by a grant from the Fund for the Improvement of Post-Secondary Education.